\documentclass[german,english]{bvm2020}

\def\articlenumber{2769}

\date{}
\title{Deep autofocus with cone-beam CT consistency constraint}

\author{Alexander~Preuhs$^1$ \and Michael~Manhart$^2$ \and Philipp~Roser$^1$ \and Bernhard~Stimpel$^1$ \and Christopher~Syben$^1$ \and Marios~Psychogios$^3$ \and Markus~Kowarschik$^2$ \and Andreas~Maier$^1$}

\authorrunning{Preuhs et al.}

\institute{%
$^1$Pattern Recognition Lab, Friedrich-Alexander-Universit{\"a}t Erlangen-N{\"u}rnberg\\
$^2$Siemens Healthcare GmbH, Forchheim, Germany\\
$^3$Department of Neuroradiology, University Hospital Basel, Switzerland
}
\email{alexander.preuhs@fau.de}

\DeclareMathOperator*{\argmin}{argmin}

\begin{document}

%
\selectlanguage{english}
\maketitle

\begin{abstract}
High quality reconstruction with interventional C-arm \textit{cone-beam computed tomography} (CBCT) requires exact geometry information. If the geometry information is corrupted, e.\,g., by unexpected patient or system movement, the measured signal is misplaced in the backprojection operation. 
With prolonged acquisition times of interventional C-arm CBCT the likelihood of rigid patient motion increases. To adapt the backprojection operation accordingly, a motion estimation strategy is necessary. Recently, a novel learning-based approach was proposed, capable of compensating motions within the acquisition plane. We extend this method by a CBCT consistency constraint, which was proven to be efficient for motions perpendicular to the acquisition plane. 
By the synergistic combination of these two measures, in and out-plane motion is well detectable, achieving an average artifact suppression of 93\,\%. This outperforms the entropy-based state-of-the-art autofocus measure which achieves on average an artifact suppression of 54\,\%.

\end{abstract}

\section{Introduction}
\textit{ Cone-beam computed tomography} (CBCT) using interventional C-arm systems has gained strong interest since an update of the guidelines of the American Stroke Association favoring mechanical thrombectomy  \cite{powers20152015,berkhemer2015randomized}. 
 The procedure needs to be guided by an interventional C-arm system capable of 3-D imaging with soft tissue image quality comparable to helical CT \cite{leyhe2017latest}. 
 This allows to perform diagnostic stroke imaging before therapy directly on the C-arm system without prior patient transfers to CT or MRI. This one-stop procedure improves the time-to-therapy \cite{psychogios2017one}, but 3-D image acquisition is challenging due to the prolonged acquisition time compared to helical CT. 
 Rigid patient head motion is more likely to occur, which leads to motion artifacts in the reconstructed slice images. 
 Thus, a robust patient motion compensation technique is highly demanded.

 Rigid patient motion in CBCT can be compensated by adapting the projection matrices, which represent the acquisition trajectory.
 This compensated trajectory is denoted as the motion free trajectory. 
 Multiple methods for rigid motion estimation in transmission imaging have been proposed, which can be clustered in three categories: (1) image-based autofocus \cite{sisniega2017motion,wicklein2012image}, (2) registration-based \cite{Ouadah2016} and (3) consistency-based \cite{Frysch2015,preuhs2019symmetry}. 
 Within those categories, learning-based approaches have been presented that detect anatomical landmarks for registration \cite{bier2018detecting,Bier2018MICCAI} or assess the reconstruction quality to guide an image-based autofocus \cite{preuhs2019image}. 
 The latter approach demonstrates promising initial results, capable of competing with the state of the art, but the motion estimation is restricted to in-plane motion \cite{preuhs2019image}.
 
As a counterpart, consistency-based methods are merely sensitive to in-plane motion, as they evaluate their consistency by the comparison of epipolar lines. For circular trajectories, epipolar lines are dominantly parallel to the acquisition plane allowing precise detection of out-plane motion. This pose consistency conditions a synergetic constraint for the deep autofocus approach presented in Preuhs et\,al.\,\cite{preuhs2019image}.

 We propose an extension of this learning-based autofocus approach which is constrained by the \textit{epipolar consistency conditions} (ECC) derived from Grangeat's theorem \cite{Aichert2015}.

\section{Motion estimation and compensation framework}
\subsection{Autofocus}

\begin{figure}[tb]
	\centering
	\includegraphics[width=\linewidth]{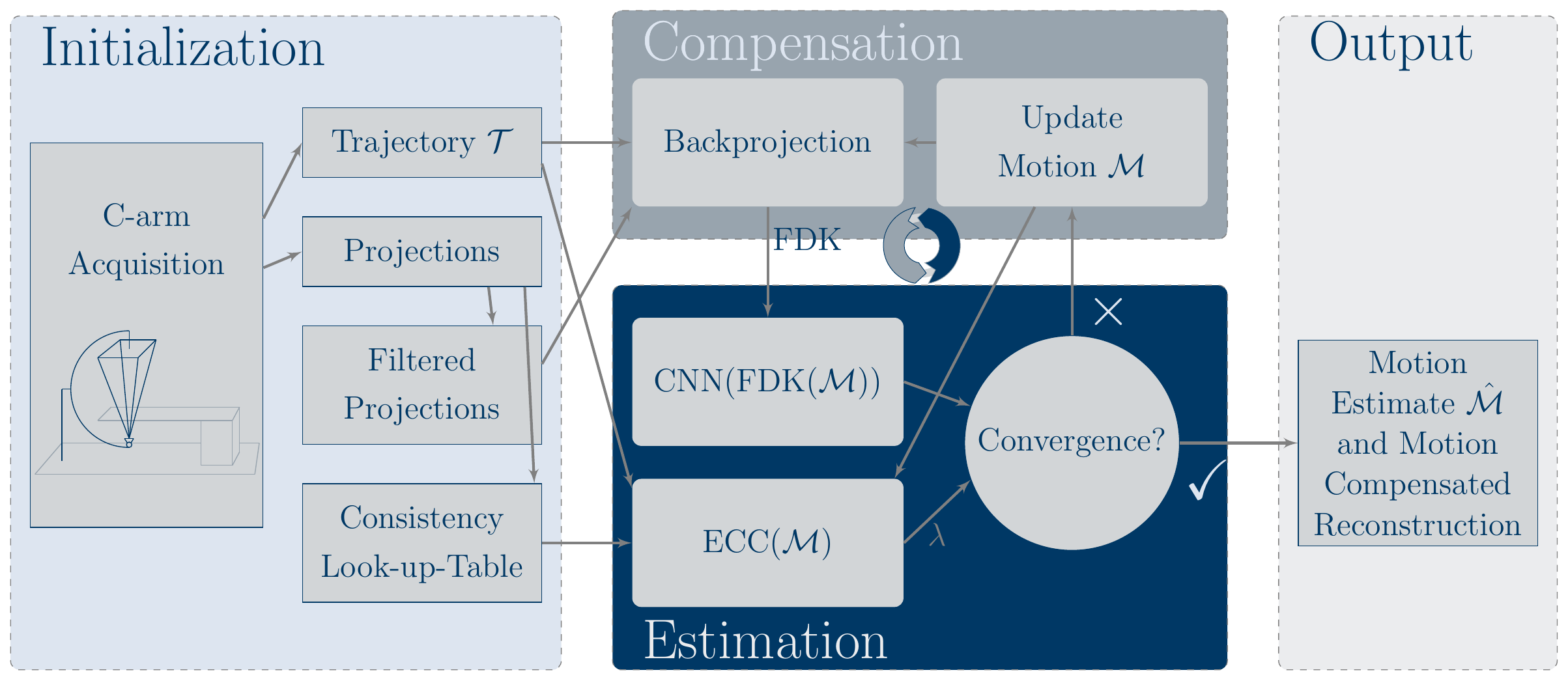}
	\caption{Flowchart describing the proposed autofocus framework. First, in the initialization, all necessary inputs for the respective methods are computed, then in an iterative compensation-estimation step the motion trajectory is derived. After convergence, the method provides the motion compensated reconstruction.}
	\label{fig:autofocus}
\end{figure}

Autofocus frameworks iteratively find a motion trajectory $\mathcal{M}$ by optimizing an \textit{image-quality metric} (IQM) evaluated on intermediate reconstructions which are updated according to the current estimated motion trajectory. The motion trajectory defines a transformation for each acquired projection $i$ representing the view-dependent patient orientation  $\vec{M}_i \in \mathbb{SE}(3)$. $\mathcal{M}$ is used, together with the offline-calibrated trajectory $\mathcal{T}$, for the backprojection operation in the \textit{Feldkamp-Davis-Kress} (FDK) reconstruction algorithm \cite{feldkamp1984practical}. If the image-quality metric saturates, the method outputs a motion compensated reconstruction as illustrated in Fig.\,\ref{fig:autofocus}.

 We use a data driven IQM which is computed by a \textit{convolutional neural network} (CNN) trained to regress the \textit{reprojection error} (RPE) from the observable motion artifacts within a reconstructed slice image. To account for out-plane motion, the autofocus framework is further constrained using the ECC based on Grangeat's theorem \cite{Aichert2015,Defrise1994}. Thus, in inference, we estimate the motion free trajectory $\hat{\mathcal{M}}$ by iteratively minimizing 
\begin{equation}
\hat{\mathcal{M}} = \argmin_\mathcal{M}   \enspace\text{CNN}(\text{FDK}(\mathcal{M})) + \lambda \cdot \text{ECC}(\mathcal{M}) \enspace,
\label{eq:IQM}
\end{equation}
with CNN(FDK($\mathcal{M}$)) denoting the network output (Sec.\,\ref{sec:regression}) for an intermediate reconstruction and ECC($\mathcal{M}$) the consistency constraint (Sec.\,\ref{sec:consistency}), both for a current motion estimate $\mathcal{M}$. The regularization weight $\lambda$ is choosen such that both metrics are within the same range.

\begin{figure}
\includegraphics[width=\linewidth]{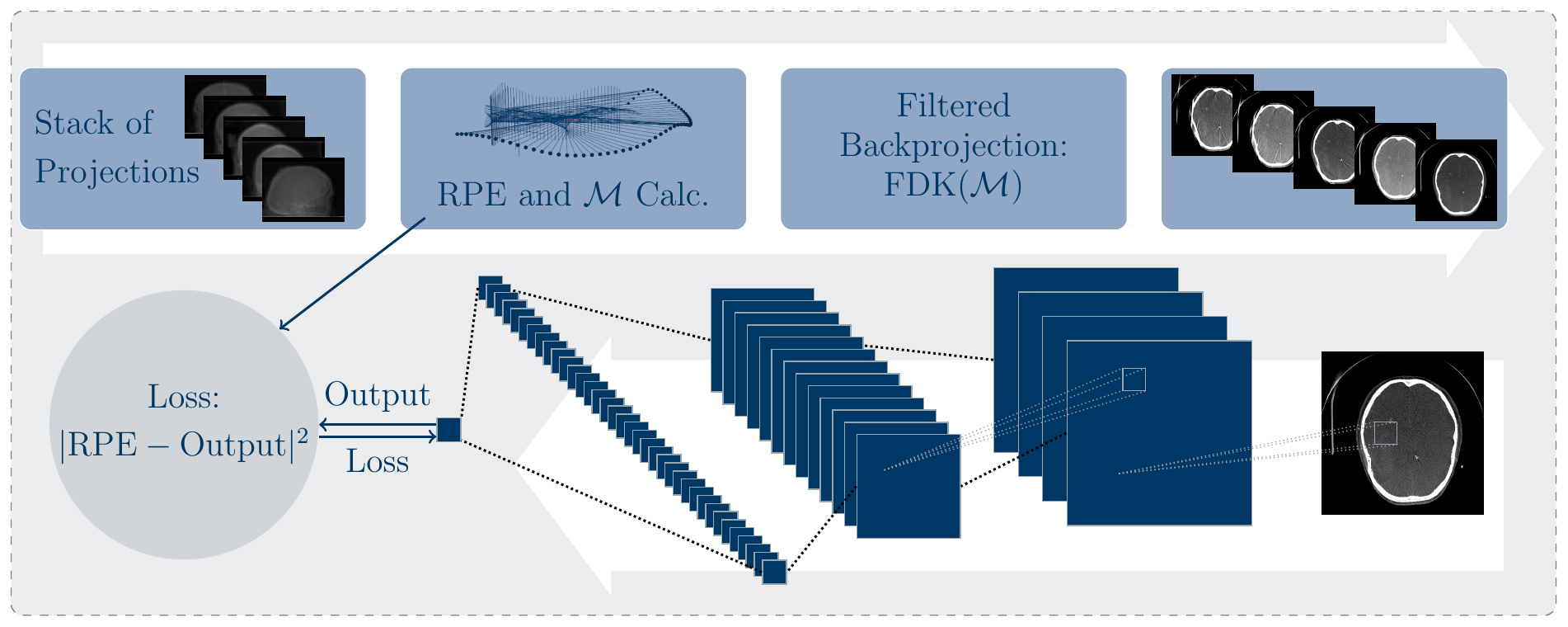}
	\caption{Schematic description of the training process for the regression network.}
	\label{fig:training}
\end{figure}
\subsection{Reprojection error regression}
\label{sec:regression}
A highly accurate method for geometry alignment is the minimization of the RPE which is computed as
\begin{equation}
\text{RPE} =  ||\vec{T}'\vec{x} - \vec{T}\vec{x}|| = ||\vec{T}\vec{M}\vec{x} - \vec{T}\vec{x}|| \enspace.
\end{equation}
This measure evaluates the reconstruction relevant deviations of projection $\vec{T}'$ and $\vec{T}$. 
When applying this measure in practice, the \mbox{3-D} position of the marker $\vec{x}$ must be determined, which is a non-trivial task, especially in the presence of geometry misalignment. 
However, the measure has nice properties, e.\,g., it increases linearly for translational motion in $\vec{M}$. 
To this end, we train a network to regress the RPE based on a reconstructed slice as depicted in Fig.\,\ref{fig:training}. 
Therefore, we calculate the RPE between a calibrated motion free trajectory and a simulated motion trajectory using virtual marker positions $\vec{x}$.

We use the same ResNet-like regression network as presented in Preuhs et\,al.\,\cite{Preuhs2018} using the same training strategy based on 16 different patients. This results in 7200 reconstructions with random motion and corresponding RPE, differing in their shape and amplitude.
\subsection{Cone-beam consistency constraint}
\label{sec:consistency}
Grangeat described a connection of the derivative in the line-integral space~---~the \mbox{2-D} Radon domain~---~of a cone-beam projection to the derivative in the plane-integral space~---~the \mbox{3-D} Radon domain \cite{Aichert2015,Defrise1994}. As a result, this measure can be used to judge the consistency of any two epipolar lines in a pair of projection images. 
The summed pairwise consistency of all sampled epipolar lines defines the consistency of a whole trajectory which we have denoted in Eq.\,\eqref{eq:IQM} as ECC. This consistency measure has been successfully applied for the compensation of motion perpendicular to the trajectory plane (out-plane) \cite{Frysch2015}. 
However, motion within the trajectory plane (in-plane) is merely detectable \cite{Frysch2015}. This poses the consistency condition improper as a cost function in a general motion compensation framework, but makes it a perfect fit as a constraint for the proposed framework. 
We expect the IQM-based motion estimation to be more robust for in-plane motion detection, because these motions distribute artifacts in axial slices. 
In contrast, out-plane motion redistributes the intensities in sagittal and coronal slices, not available to the network.

\section{Experiments and results}


\def\arrvline{\hfil\kern\arraycolsep\vline\kern-\arraycolsep\hfilneg}
\begin{table}
	\centering
	\caption{Motion compensation results for the proposed autofocus method (Proposed) and a state-of-the-art autofocus method employing entropy as an IQM (Entropy) using in-plane and out-plane motion applied to the validation and test patients. Artifact suppression describes the improvement of the RMSE, with 0\% being no improvement w.r.t. the uncompensated reconstruction (No Comp.) and 100\% denoting a complete recovery of the ground truth reconstruction.}
	\label{tab:motion_eval}
	
	\resizebox{\textwidth}{!}{
		{\def\arraystretch{1.1}\tabcolsep=1pt
			\begin{tabular}{@{\extracolsep{10pt}} ll c c c   c c @{}}\hline\hline
				&\multirow{2}{*}{Dataset} & \multicolumn{2}{c}{Artifact Suppression}&\multicolumn{3}{c}{SSIM}\\ 
				\cline{3-4} \cline{5-7}
				&	   		&Entropy & Proposed &No Comp. &Entropy &Proposed \\
				\hline
				\multirow{3}{*}{In-Plane}	 &	Val. 1 		&   77.0\,\%  & \textbf{94.2}\,\%   &  0.814 & 0.892  &\textbf{ 0.944  }  \\
				&	Val. 2 		&   89.3\,\%  & \textbf{96.5}\,\%   &  0.742 & 0.890  &\textbf{ 0.918 }    \\
				&	Test 1 		&   21.4\,\%  & \textbf{90.1}\,\%   &  0.835 & 0.855  &\textbf{ 0.931 }    \\\hline
				\multirow{3}{*}{Out-Plane} &		Val. 1 	&   92.1\,\%  & \textbf{96.8}\,\%   &  0.911 & 0.963  & \textbf{0.983  }   \\
				&	Val. 2 		&   44.4\,\%  & \textbf{86.8}\,\%   &  0.926 & 0.939  &\textbf{ 0.981 }   \\
				&	Test 1 	    &   2.0\,\%  & \textbf{94.0}\,\% &  0.904 & 0.901  & \textbf{0.990 }      \\\hline 
				\hline 
				
			\end{tabular}
	}}
\end{table}


We generate motion trajectories using Akima splines. To prevent an inverse crime scenario we use a Piecewise Cubic Hermite Interpolating Polynomial (PCHIP) for the estimate of the motion trajectory. 
 By modeling the motion with a spline, we achieve a lower computational complexity in the optimization, as only the spline nodes need to be estimated instead of every motion matrix and additionally we implicitly regularize the motion  to be smooth. 
 The motion compensation is performed by optimizing Eq.\,\eqref{eq:IQM} using the Nelder-Mead simplex method. We initialize the simplex corresponding to the initial estimated RPE. 
  If the initial RPE is close to zero, we expect only a small motion, whereas a high initial RPE is expected for greater motion amplitudes. We use a block-based motion estimation strategy,  where within each block we only optimize  a subset of neighboring spline nodes, while keeping the other nodes fixed. To enforce $\vec{M}_i \in \mathbb{SE}(3)$ we construct the motion matrices as a function of rotation and translation parameters.

\begin{figure}
	\begin{center}
		\resizebox{1\textwidth}{!}{
			{\def\arraystretch{1}\tabcolsep=2pt 
				\begin{tabular}{ccccccc}
\includegraphics[]{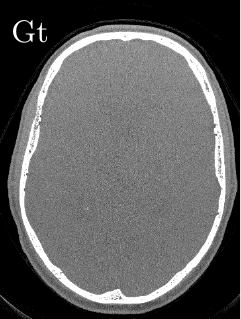}
					&
\includegraphics[]{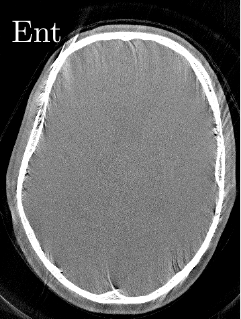}
					&
\includegraphics[]{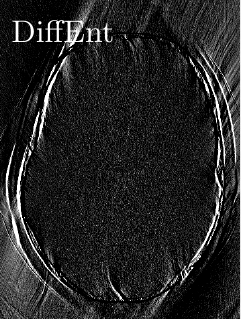}
					&{\color{white} a}&
\includegraphics[]{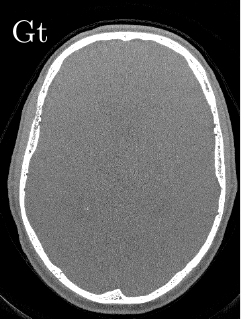}
					&
\includegraphics[]{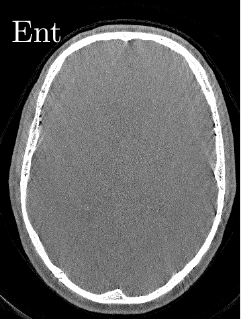}
					&
\includegraphics[]{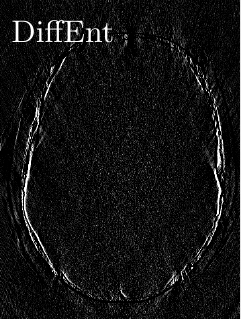}
					\\
\includegraphics[]{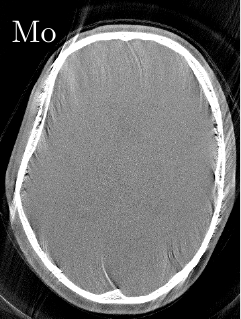}
					&
\includegraphics[]{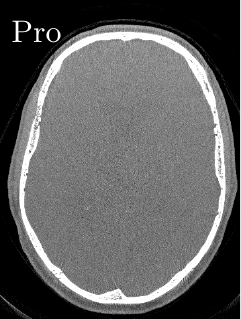}
					&
\includegraphics[]{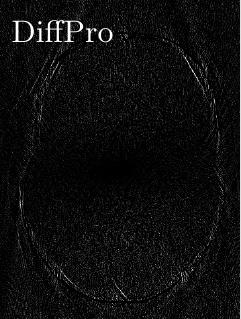}
					&&
\includegraphics[]{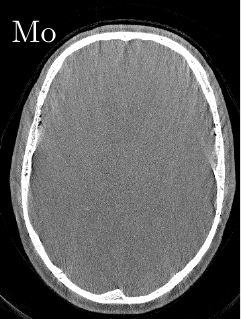}
					&
\includegraphics[]{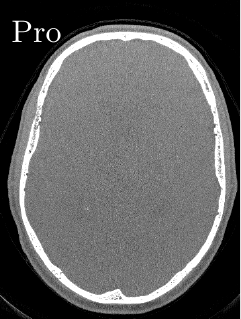}
					&
\includegraphics[]{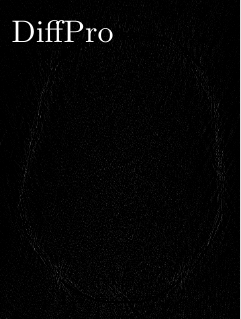}				
					\\
					\multicolumn{3}{c}{In-Plane Motion}&&
					\multicolumn{3}{c}{Out-Plane Motion}
				\end{tabular}%
		}}
	\end{center}
	\caption{Reconstructions of the test patient visualized using a bone-window. For in-plane and out-plane motion, the ground truth (Gt) together with the motion affected reconstruction (Mo) is depicted in the respective leftmost columns. Next to it, the motion compensation result using the state-of-the-art entropy-based autofocus (Ent) and the result using our proposed method (Pro) are depicted. The respective rightmost column shows the difference image of the  compensated images to the ground truth using entropy (DiffEnt) or the proposed method (DiffPro), respectively.}
	\label{fig:motion_reco}
\end{figure}

Using the motion affected trajectory we perform an initial reconstruction and apply the proposed motion compensation algorithm (Fig.\,\ref{fig:autofocus}). In addition, we use the best performing IQM from \cite{wicklein2012image} as a baseline compensation. Quantitative results for the test patient and the two validation patients are described in Tab.\,\ref{tab:motion_eval} and visual results for the test patient are depicted in Fig.\,\ref{fig:motion_reco}.


\section{Conclusion and discussion}
We present a novel combination of synergistic motion estimation strategies: a consistency-based method for out-plane motion estimation and a novel data-driven image quality metric for the estimation of in-plane motion. The presented metric is computed fast, as the network relies only on an axial slice, reducing the computational burden of the FDK to that of a fan-beam reconstruction. In addition, the consistency-based metric is inherently fast, because the consistency look-up-table can be pre-computed (Fig.\,\ref{fig:autofocus}) and remains constant during the motion compensation. The final motion compensated reconstruction is solely based on the projection raw-data and the estimated motion trajectory. This allows us to use a learning-based approach while ensuring data integrity, as we are not manipulating any of the raw data.   
\\{}\\~%
\textbf{Disclaimer:} The concepts and  presented in this paper are based on
research and are not commercially available.

\bibliographystyle{bvm2020}

\bibliography{0000}
\marginpar{\color{white}E\articlenumber} 
\end{document}